# Research on gesture recognition method based on SEDCNN-SVM

Mingjin Zhang [a], Jiahao Wang [b], Jianming Wang [c, d], Qi Wang [a, e*]

[a] School of Electronic and Information Engineering, Tiangong University, Tianjin 300387, China

[b] College of arts and science, Boston University, Boston 02215, the United States

[c] School of Computer Science and Technology, Tiangong University, Tianjin 300387, China

[d] Tianjin Key Laboratory of Autonomous Intelligence Technology and Systems, Tiangong University, Tianjin 300387, China

[e] Tianjin Key Laboratory of Optoelectronic Detection Technology and System, Tiangong University, Tian 300387, China

**A B S T R A C T**

Gesture recognition based on surface electromyographic signal (sEMG) is one of the most used methods. The traditional manual feature extraction can only extract some low-level signal features, this causes poor classifier performance and low recognition accuracy when dealing with some complex signals. A recognition method, namely SEDCNN-SVM, is proposed to recognize sEMG of different gestures. SEDCNN-SVM consists of an improved deep convolutional neural network (DCNN) and a support vector machine (SVM). The DCNN can automatically extract and learn the feature information of sEMG through the convolution operation of the convolutional layer, so that it can capture the complex and high-level features in the data. The Squeeze and Excitation Networks (SE-Net) and the residual module were added to the model, so that the feature representation of each channel could be improved, the loss of feature information in convolutional operations was reduced, useful feature information was captured, and the problem of network gradient vanishing was eased. The SVM can improve the generalization ability and classification accuracy of the model by constructing an optimal hyperplane of the feature space. Hence, the SVM was used to replace the full connection layer and the Softmax function layer of the DCNN, the use of a suitable kernel function in SVM can improve the model's generalization ability and classification accuracy. To verify the effectiveness of the proposed classification algorithm, this method is analyzed and compared with other comparative classification methods. The recognition accuracy of SEDCNN-SVM can reach 0.955, it is significantly improved compared with other classification methods, the SEDCNN-SVM model is recognized online in real time.

# 1. Introduction

Surface electromyographic signal (sEMG) is a signal associated with muscle contraction, which can reflect neuromuscular activity to a certain extent [1]. It has the advantages of non-invasive, non-invasive, and simple operation [2]. Therefore, sEMG has perfect applications and prospects in prosthetic limb control [3], it also used for clinical diagnosis and neurological rehabilitation [4].

Classification and recognition of different gestures by sEMG are important in medical research [6]. The naturalness and flexibility of human-computer interaction are affected by development of gesture recognition [5]. In recent years, people can control the movement of robots by collecting sEMG of different gestures, so that robots can make corresponding gestures. The application includes the control of rehabilitation prosthesis and rehabilitation exoskeleton [7][8]. People with inconvenient hand movements or patients with hand disabilities can use prosthetic limbs to carry out some simple hand rehabilitation training or bring some convenience to the daily life [9].

At present, many pattern recognition techniques are applied to classify and recognize sEMG [11]. When classifier methods are used, it is important to extract the features of the collected data. Common features in sEMG classification are based on time spectrum analysis [10], which can be divided into time domain features and frequency domain features [12-13]. For example, Hussein F. Hassan et al. extracted six-time domain features from each segment of sectioned sEMG, such as the mean value (MAV), root mean square (RMS) and zero cross (ZC). Seven hand motions were classified using LDA, KNN, and SVM classifiers [3]. Compared with other machine learning classifiers, the SVM classifier has a better classification effect and more stable classification performance. This is because SVM can improve the generalization ability of the classifier by maximizing the bounding when dealing with complex data compared to KNN and LDA classifiers. José Jair A. Mendes Junior used the wrapper-by-feature selection approach with methods of feature dimensionality reduction, three frequency domain features were extracted from eight-channel sEMG, such as median frequency (MF), mean power frequency (MPF) and maximum power spectrum (MPS). A high recognition accuracy could be achieved by using three frequency domain features combined with a significant boundary nearest neighbor dimensionality reduction method and the SVM [14]. The extracted feature information could be trained and classified by the SVM; the generalization ability of the model could be improved by maximize bounding.

Since the traditional manual feature extraction can only extract some low-level signal features. Deep learning can be seen as a machine learning technique based on multi-level neural networks, with the aim of learning feature representations from complex data, so that the classification tasks can be achieved. Deep learning can automatically learn feature representations from raw data, without manually designing features, the complex structure of data can be captured, and it supports the end-to-end learning process. Deep learning can further enhance the model's ability to express data by increasing the depth of the network. Complex nonlinear relationships can be processed more effectively by deep network models [15]. Jiang et al. used a deep convolutional neural network (DCNN) algorithm to process the sEMG of shoulder and upper arm muscles from upper limb movement to identify the corresponding movement. They compared the recognition accuracy of different CNN models [20]. It could be concluded that with the deepening of the CNN network, there was a gradient disappearance, it affects network's extraction characteristics and classification performance, so that it resulted in the relatively low classification accuracy of the network.

Based on the above research, a recognition method based on the combination of DCNN and SVM (SEDCNN-SVM) is proposed, the data feature vectors extracted by DCNN is used as input to SVM. The contributions of this study are as follows:
1) The DCNN-SVM network model was built. The superiority of DCNN in extracting feature information and the stability of SVM in handling complex data classification tasks were used by DCNN-SVM. The classification method can handle complex data classification problems and significantly improve classification ability of the model.
2) SE-Net and residual module are added to the model to improve the model and solve the problem of missing essential signal features in the process of convolution.
3) The validity of the SEDCNN-SVM model is validated by the sEMG dataset of ten subjects with ten different gestures. The SEDCNN-SVM model is recognized online in real time.

The remaining structure of the paper is as follows: Section 2 introduces the equipment of sEMG acquisition, the process of sEMG acquisition, the pretreatment of sEMG; In section 3, the SEDCNN-SVM method is introduced and analyzed. Section 4 provides the experiment results and analysis.

## 2. Materials and Methods

The data set of this study was obtained by collecting sEMG of different gestures with Myo armband. The Myo armband was placed on the right forearm of ten different subjects. sEMG was processed by filtering, sliding window segmentation and feature extraction [17]. Gesture classification was performed using the SEDCNN-SVM classification method. The flow chart is shown in Figure. 1.

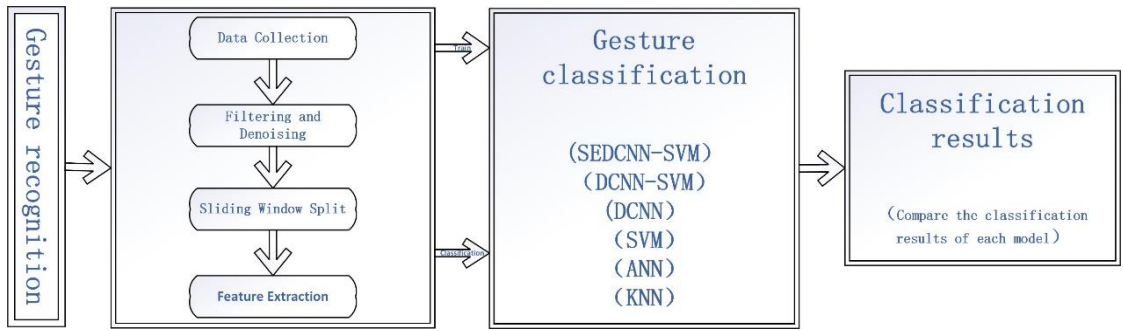

**Fig. 1.** Main methods of research.

2.1 Data Collection

To conveniently collect sEMG of hand movements, the Myo armband was employed in this study as the acquisition device [16-17]. Myo armband is a low-cost, high-performance, wearable device [3]. It represents an innovative armband launched by Thalmic Labs, a Canadian start-up company with 8 sEMG channels, as shown in Figure. 2 (a). When collecting sEMG, the Myo armband was worn on the subject's right forearm, as shown in Figure. 2(b). The sampling frequency of sEMG was 200Hz.

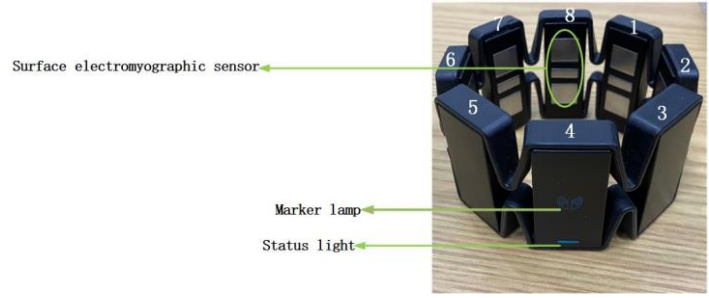

(a)

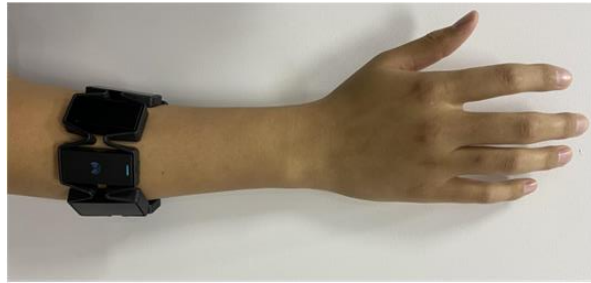

(b)

**Fig. 2.** Data collection: (a) Myo armband (b) Myo armband placed on the right forearm.

The data were collected from ten healthy subjects, including five males and five females, aged between twenty and thirty. The details of the subjects are shown in Table 1. Before the experiments, participants were informed and filled out a written informed consent form.

Table 1. The main information in the dataset

| Number of subjects | Male to female ratio | Average height(cm) | Average weight(kg) | Average BMI(kg/m$^2$) |
|---|---|---|---|---|
| 10 | 1:1 | 170.5±9.2 | 60.17±3.12 | 21.35±3.18 |

In the process of sEMG signal data collection, ten different hand movements are planned: fist, one, two, OK, open hand, praise, six, up, down, eight, as shown in Figure. 3. The subject worn the Myo armband on the right forearm and straighten the arms perpendicular to the ground. The subjects were trained on the corresponding hand movements before the formal experiment. The subjects were told not to do strenuous exercise before the experiment to avoid the effects of muscle fatigue during the experiment [17].

Before the signal collection, skin hair at the collection site of the subject was removed. 75% alcohol cotton was applied to wipe the skin surface to remove the affected oil and dead skin, so that good contact between the skin of the subject and the electrode could be ensured. Effective information of muscle activity could be obtained from the pure sEMG collection [17-20].

While collecting the EMG signals of hand movements, the subjects were asked to repeat 15 times, each movement was recorded for 16s. After each gesture, the subjects were given five seconds to relax their hands, this was to reduce fatigue in their hand muscles. In this experiment, 3200 sampling points were collected for each hand movement.

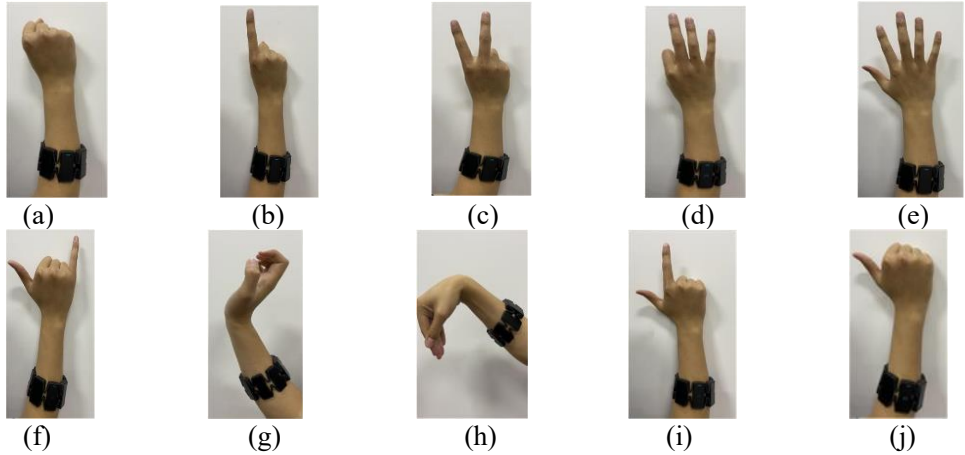

**Fig. 3.** Classified gestures:(a) fist (b) one (c) two (d)OK (e) open hand (f) six (g) up (h) down (i) eight (j) praise.

2.2 Filtering and Denoising

When collecting sEMG of different gestures, some redundant noise signals are mixed [17]. Therefore, sEMG should be preprocessed by filtering and denoising first. Effective information on sEMG collected in this study is mainly distributed between 20Hz and 200Hz [21]. SEMG is interfered with power frequency signals of 50Hz and low-frequency signals of less than 20Hz. In this paper, a Butterworth high-pass filter and a trap filter are selected to filter out low-frequency noise below 20 Hz and industrial frequency interference at 50 Hz.

The design of the Butterworth filter is simple, it is easy to make, so it has been widely used [22]. A Butterworth low-pass filter can be expressed by formula (1), where $n$ is the order of the filter and $w_c$ is the cutoff frequency, the value of $w_c$ is 20, and the value of $n$ is 8.

$$|H(w)|^2 = \frac{1}{1+\left(\frac{w}{w_c}\right)^{2n}} \quad (1)$$

A trap filter can block the passage of signals of a specific frequency. An ideal filter can be expressed by formula (2), with $w_0$ = 50Hz, which can filter out 50Hz power-frequency noise.

$$|H(e^{jw})| = \begin{cases} 1, & w \neq w_0 \\ 0, & w = w_0 \end{cases} \quad (2)$$

After two stages of filtering, the spectrum diagram before and after sEMG filtering is drawn, as shown in Figure. 4. By comparison, it can be concluded that sEMG collected by the Myo armband is relatively stable [8].

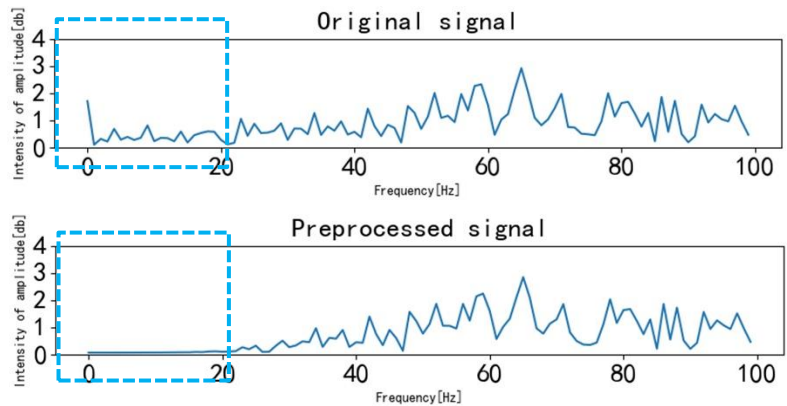

**Fig. 4.** sEMG before and after filtering.

2.3 Sliding Window Split

To facilitate data calculation, filtered sEMG was segmented by the method of sliding time window. Several sEMG data segments were obtained for signal characteristics calculation and analysis [17-19]. The calculation formula for the number of overlapping windows is shown in Formula (3), *W* is the number of Windows, *n* is the number of sampling points of sEMG, *a* is the size of the window, and *b* is the sliding step size [23]. In this study, the sliding window and sliding step size were set to 200 and 100, respectively. Fig. 5 shows a schematic diagram of sliding window segmentation.

$$W = \frac{n-k}{s} + 1 \tag{3}$$

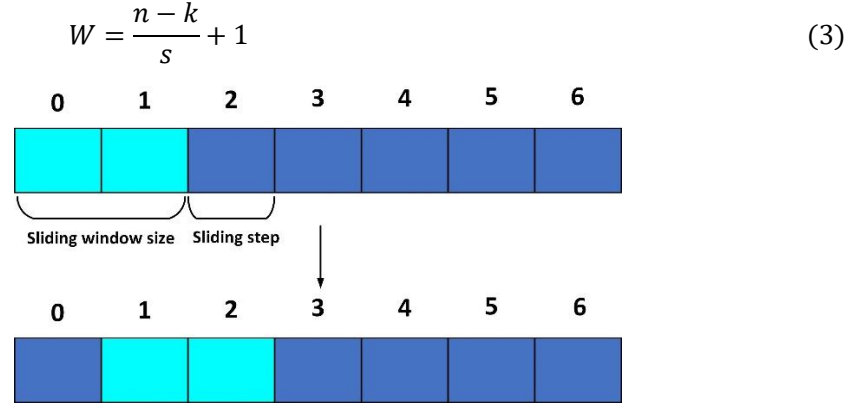

**Fig. 5.** Schematic diagram of sliding window segmentation.

2.4 Feature Extraction

Feature extraction is the most critical step in gesture recognition and classification. Extracting features of different types from the EMG window is conducive to improving the recognition accuracy of the classifier [24]. In this study, feature extraction of sEMG was carried out from multiple perspectives. The mean value (MAV), root mean square (RMS), standard deviation (STD), waveform length (WL), waveform amplitude (WA), zero crossing (ZC), slope sign change (SSC), and integrated electromyography (IEMG) were extracted from the time domain. The core of the frequency-domain feature extraction method is the Fourier transform (FT) [26]. In this study, fast Fourier transforms (FFT) was used to extract frequency-domain features of sEMG, the intensity and distribution of each frequency component in the signal can be clearly showed by frequency domain features, they can further extract effective information of muscle activity in sEMG data [27]. Features of average power frequency (MPF) and median frequency (MF) were extracted from the frequency domain [25]. The extracted feature information is used for the machine learning classifier in the comparison method.

**3. Gesture Recognition Methods**

In this study, the classification model of SEDCNN-SVM is proposed for the task of classifying and recognizing sEMG with different gestures. This model can extract more important feature information and improve the accuracy of classification and recognition.

3.1 Network Construction

Convolutional neural network (CNN) extracts the image features through filters. Weight sharing is a main feature of CNN, which will greatly improve the training speed [17]. It is a kind of feedforward neural network, and its characteristic is that the neuron nodes of each layer only respond to the neurons within the local area of the previous layer [27][28]. Deep convolutional neural network model (DCNN) is generally composed of several convolutional layers superimposed on several fully connected layers, which are to deepen the general convolutional neural network model.

In this study, DCNN network was used to extract sEMG features by convolution operation, SVM classifier was used to identify and classify. The full connection layer and Softmax activation layer of DCNN network were replaced by SVM, the global average pooling layer was used as the last layer of DCNN to play the role of tiling. For traditional CNNs, simply increasing the depth to improve network performance will lead to degradation. To overcome degradation, so SE-Net and the residual block were added to the network model, so that the DCNN network model was well optimized. The residual structure can connect the initial input information and the output characteristic data of the entire DCNN network. Figure. 6 shows the architecture of the network model.

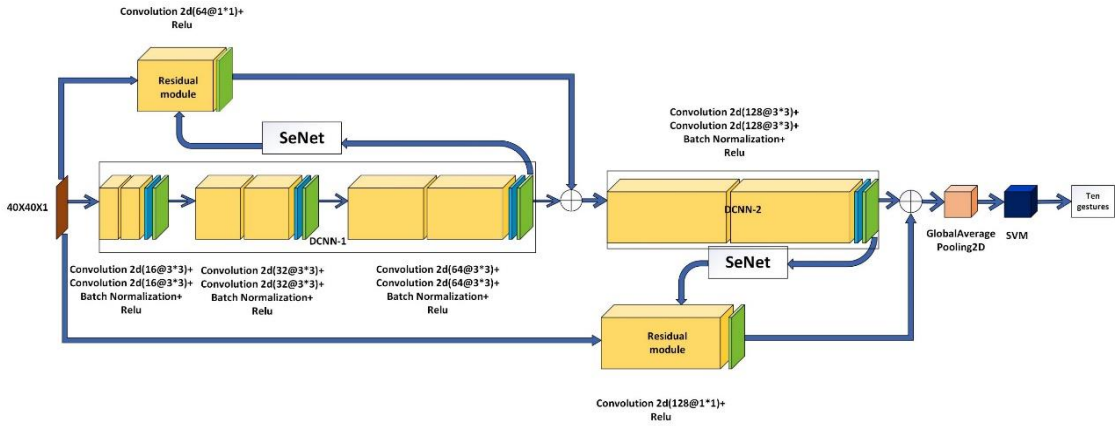

**Fig. 6.** Architecture of the network model.

Two subblocks are used by the DCNN network to extract signal features: DCNN-1 and DCNN-2. The DCNN-1 block contains three convolutional layers, the DCNN-2 block contains one convolutional layer. There is a batch normalized BN layer between each convolutional layer and the activation layer.

A $40 \times 40 \times 1$ matrix data is input into the network; four deep convolution modules are designed to automatically extract the features of the data. Each convolution module is composed of two two-dimensional convolution kernels, one BN layer and one activation layer. The convolution filter size is $3 \times 3$. The four convolution layers that are used to perform convolution operations, the number of their convolution kernel is 16, 32, 64 and 128, respectively. SE-Net can be easily integrated with DCNN network to improve the performance of DCNN network. The residual module contains only one filter with dimensions $1 \times 1$, the number of convolution kernels for the residual structures are 64 and 128, respectively. The residual structure can be the initial input information of the whole DCNN network and the characteristics of the output data, it also can effectively alleviate the problem of network degradation, retain the original information; Feature vectors are tiled and expanded by a global average pooling layer. Feature vectors of extracted sEMG data are sent to SVM classifier for recognition and classification.

3.2 Support and Vector Machine (SVM)

SVM is an effective recognition method in sEMG of different gestures. The basic idea is to solve the separation hyperplane, which can correctly divide the training data set and have the largest geometric interval to obtain the optimal classification results [29]. SVM performance would be significantly determined by the kernel function parameters. Hence, the parameters of SVM should be appropriately set to satisfy a high modelling capability under various identification situations. The Gaussian kernel function is used as the kernel function of SVM in this study, its expression is shown in Formula (4). The Gaussian kernel function enables SVM to deal with nonlinear classification problems of complex data, which can help to improve the generalization ability and the classification stability of SVM.

$$K(x_1, x_2) = exp\left(-\frac{\|x_1-x_2\|^2}{2\sigma^2}\right) \quad (4)$$

3.3 Squeeze and Excitation Network (SE-Net)

The deep-learning attention mechanism is derived from the human visual system, which can help the model assign different weights to each part of the input. For the task of sEMG recognition and classifier, the attention weight matrix directly reflects the area of interest for the model [25].

SE-Net is a new network architecture based on CNN, which can be inserted into different types of networks. The network structure of SE-Net is shown in Figure. 7. SE-Net can extract the feature information of the region of important corresponding to the sEMG distribution, so that the interpretability of the model could be improved. The input is characteristic of a $H \times W \times C$, where $C$ is the number of channels. Then an output of $1 \times 1 \times C$ scalar is obtained by a global average pooling of each channel. Two convolution layers are used as the full connection layer in this study, they are employed in the model to capture the channel dependencies of features from the statistics obtained by the global average pooling. The output is activated by the sigmoid function to obtain the channel dependencies. The new weights are assigned to the characteristics of the C channels. The obtained weight is combined with the characteristics of input data to obtain the data characteristics of output (each element of the corresponding channel is multiplied by the weight of each channel).

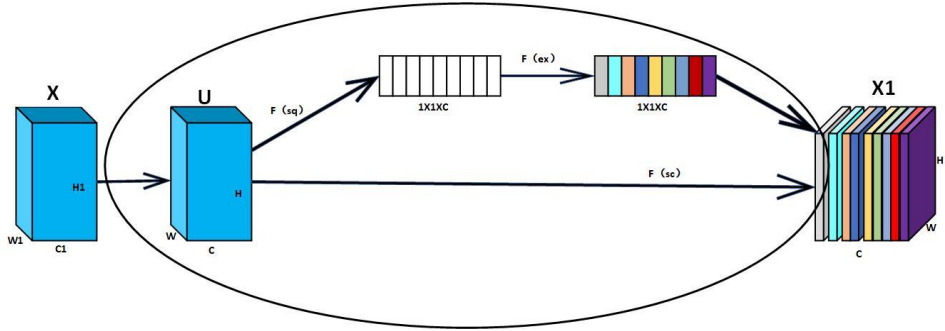

**Fig. 7.** SE-Net network structure.

The regression vectors created by each forest are constructed using K-fold cross-validation to limit the danger of overfitting [38]. The model structure of the DF is shown in Figure 9.

3.4 Residual Module

The nonlocal residual block is obtained through mapping the whole features to the features error, a further step toward residual learning is achieved by the network [17]. The input feature is directly connected to the output feature result from a subsequent layer by skipping connection. The structure diagram of the residual module is shown in Figure 8. Due to the introducing of residual block, the original feature information is fused with the residual feature information, which makes the network training more efficient and stable [26].

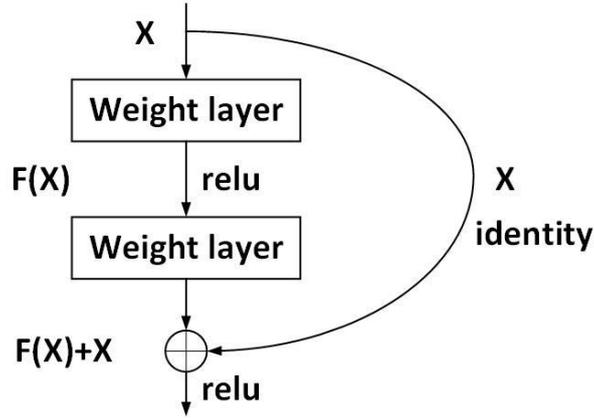

**Fig. 8.** Residual module structure.

In general, the classification method proposed in this paper has the following advantages:

1. The advantages of signal feature extraction by DCNN network and the stability of signal classification by SVM classifier are fully used by SEDCNN-SVM. The model significantly improves the classification ability and the generalization ability of the model.

2. SE-Net and residual modules are added to the substructures of network model DCNN-1 and DCNN-2 respectively to improve the model. By combining the original information of the signal with the feature information, the problem of feature loss in the convolution during feature extraction for the DCNN network is solved, so that the expression of the important features of the signal is preserved. As a result, the performance of the network is optimized.

## 4. Results and Discussion

As shown in Figure. 3, Butterworth high-pass filter and trap filter are used to filter 3200 sEMG data signals from ten different gestures. All filtered signals are segmented by the sliding window method. Eight-time domain features and two frequency domain features are extracted. Of all the data, 80% data is for training and the remaining 20% data is for testing and validation. The network architecture of DCNN is realized by Tensorflow and Kreas. Adam is used as the optimization algorithm for propagation in the network model, inter class cross entropy is used as the loss function, random inactivation is set to 0.5, the learning rate is set to 0.001, the number of iterative training and verification is set to 100. The data of different gestures were put into the model for classification and recognition, the results of classification and recognition were obtained. Several evaluation indexes of classification and recognition were obtained to verify the accuracy of the model classification and recognition. By comparing the recognition results of different algorithms, it is proved that the algorithm proposed in this study has higher recognition accuracy for gestures.

The sEMG signal data of ten different gestures were sent into the SEDCNN network model for training and verification. The training results of sEMG data are shown in Figure. 9.

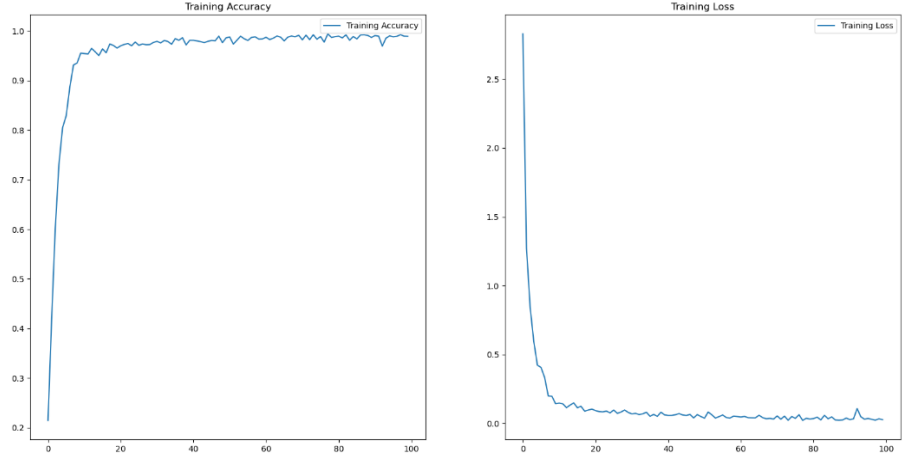

**Fig. 9.** Curve of training accuracy and loss function of SEDCNN network

Figure 9 shows an attenuation curve of accuracy and loss value in the training process. The network reaches an accuracy of approximately 90% before the 10th epoch, with a slight rise from the 10th to the 20th epoch. It achieves stable training accuracy after the 20th epoch.

A designated DCNN trained model was finalized through iterative descent to the lowest level. During the training process of SEDCNN network, the optimal network model parameters are saved. The parameters of the model are shown in Table 2.

**Table 2.** Parameters of the SEDCNN-SVM model.

| Model | Layers | Parameters |
|---|---|---|
| | Conv2d | 16@3×3 |
| | Conv2d | 16@3×3 |
| DCNN-1 | Conv2d | 32@3×3 |
| | Conv2d | 32@3×3 |
| | Conv2d | 64@3×3 |
| | Conv2d | 64@3×3 |
| DCNN-2 | Conv2d | 128@3×3 |
| | Conv2d | 128@3×3 |
| | Batch-size | 16 |
| Hyperparameter | Learn-rate | 0.001 |
| | Dropout | 0.5 |
| Residual | Conv2d | 64@3×3 |
| | Conv2d | 128@3×3 |
| SE-Net | Conv2d | channel@3×3 |

Extensive confirmatory and comparative experiments with sEMG gestures recognition is conducted and compared with DCNN-SVM, DCNN, SVM, ANN and KNN methods. Particularly, the BP neural network is used in ANN. BP neural network is a multilayer feedforward network trained by error back propagation. Accuracy, Recall, Precision and F1-score classification were obtained to compare the identification results of the models, as defined in equations (5) – (8). So that the accuracy and validity of SEDCNN-SVM classification method in sEMG gesture recognition could be demonstrated.

Equations (5) – (8) explain the metrics calculation. The accuracy is calculated by comparing the number of samples correctly classified and the total number of samples. The recall, or sometimes called sensitivity, measures how many samples of a class are properly classified by dividing the number of true positives (TP) into the number of samples in the class, which can be obtained by the sum of the TP and false negatives (FN). The precision, or sometimes called positive predicted value (PPV), gives a measure of the number of correctly predicted samples for each class, this can be calculated by dividing the number of TP into all the predicted samples of the class, which can be calculated by summing the TP and the FP. The F1-Score gives a harmonic mean between recall and precision. This is useful to avoid bias in some systems having many true negatives (TN) and whose accuracy does not represent correctly the classifier performance.

$$Accuracy = \frac{TP+TN}{TP+TN+FP+FN} \tag{5}$$

$$Precision = \frac{TP}{TP+FP} \tag{6}$$

$$Recall = \frac{TP}{TP+FN} \tag{7}$$

$$F1 = 2 \times \frac{Precision \times Recall}{Precision+Recall} \tag{8}$$

Table 3. Results of ten gestures categories.

| Classifier | Accuracy | Recall | Precision | F1-score |
|---|---|---|---|---|
| SEDCNN-SVM | 0.955 | 0.954 | 0.965 | 0.957 |
| DCNN-SVM | 0.912 | 0.912 | 0.914 | 0.912 |
| DCNN | 0.904 | 0.902 | 0.911 | 0.903 |
| SVM | 0.600 | 0.597 | 0.603 | 0.594 |
| ANN | 0.801 | 0.803 | 0.811 | 0.801 |
| KNN | 0.574 | 0.576 | 0.579 | 0.573 |

It can be seen from the experimental results that the classification method of DCNN-SVM has a higher classification accuracy than the method of DCNN network and SVM classifier alone. The features of sEMG classified by SVM classifier are eight-time domain features and two frequency domain features extracted manually. It can be found that compared with KNN classifier, SVM classifier has higher classification accuracy and more stable classification effect. The data with higher dimensions can be effectively dealt by SVM. SVM is divided by building a hyperplane that maximizes class intervals, so that the prediction is relatively fast. The Gaussian kernel function in SVM can also fit the complex data distribution and improve the generalization ability of classifier. However, the classification recognition accuracy of both SVM and KNN classifiers for different gestures of sEMG is very low, due to the limitation of manual feature extraction for the machine learning method, some low-level signal features can only be extracted. As a result, the signal features themselves lack diversity; when dealing with complex and large data sets, SVM and KNN classifiers have a high computation complexity, this result in SVM classifier is sensitive to missing data features, the space complexity of KNN classifier is high. Although the classification effect of the ANN model is better than SVM and KNN classifier, the recognition accuracy is still lower than the DCNN network model. Furthermore, the convergence speed of the ANN model is slow, and the training ability and prediction ability are poor, especially when dealing with complex and large samples. Feature extraction using DCNN network model is much better than manual extraction. Moreover, SE-Net and residual modules in the SEDCNN network can improve the data feature extraction performance in DCNN network, the gradient disappearance of network model due to the increase of network depth is avoided. Not only the problem of network degradation is solved by this method, but also the expression of signal features is enhanced. In the process of convolution, feature loss is reduced, so the integrity of signal features is guaranteed. Likewise, the importance weights of each feature channel can be learned by the model, for this reason, the model can enhance the response to important features, improve the feature representation and classification performance of the network, and maintain a high computational efficiency. In conclusion, compared with separate DCNN and SVM classification methods, SEDCNN-SVM is more convenient to deal with complex data classification problems. The feature information in sEMG signal can be captured better, the powerful feature learning ability of DCNN and the accurate classification in high-dimensional space for SVM. A higher classification accuracy can be achieved.

Some SOTA algorithms developed in the field of gesture recognition were compared with SEDCNN-SVM. The results are given as follows:

1) GCN: Different gestures are recognized based on the postural graph convolutional network (GCN) method. Related gestures of hand and body are taken as the input mode of the network, and the GCN, residual connection and residual module structure are used for recognition and classification.
2) MSMHA-VTN: The pyramid structure of multi-scale features is extracted by using the multi-scale multi-attention time-frequency converter network (MSMHA-VTN) and the multi-scale head attention model of the transformer. The model adopts different attention dimensions for each head of the transformer, so that it can provide attention mechanism at the multi-scale level.
3) Alexnet+Adam: Time-order features from sEMG signals can be extracted by Alexnet. The extracted features are classified through the full connection layer, the output is the probability distribution of each gesture. The Adam optimizer is used to adjust network weights, so that loss functions are minimized. The classification method improves the accuracy of gesture recognition.
4) CNNSP+CNN: A 12-layer CNN was built as the backbone. The improved 20-channel data enhancement method was used to avoid overfitting. It was used to recognize and classify different gestures.

Table 4. Results of ten gestures categories.

| Classifier | Accuracy | Recall | Precision | F1-score |
| --- | --- | --- | --- | --- |
| SEDCNN-SVM | 0.955 | 0.954 | 0.965 | 0.957 |
| GCN | 0.926 | 0.929 | 0.931 | 0.926 |
| MSMHA-VTN | 0.912 | 0.912 | 0.918 | 0.910 |
| Alexnet+Adam | 0.915 | 0.914 | 0.921 | 0.916 |
| CNNSP+CNN | 0.908 | 0.908 | 0.913 | 0.906 |

According to the data in Table 4, the recognition accuracy of the algorithm proposed in this paper is improved compared with SOTA algorithms, which indicates the feasibility of the new algorithm. GCN need to spend a lot of computing time when dealing with complex data, the effect of recognition and classification is limited. When the training data is too complex, the model's ability to generalize is affected. In addition, the performance of the model is easily affected by the parameters, this results in a low recognition accuracy of the model. The training time of MSMHA-VTN is long, the final performance of the model is affected by overtraining. The high complexity leads to the precision reduction. However the structure of the Alexnet network is deep, it is easy to overfit during training. Adam is easily affected by hyperparameters, and inappropriate hyperparameters leads to poor classification. Features at different scales can be captured by CNNSP, but the computational complexity is increased, so that feature representation becomes poor. The expression of feature information for each channel is enhanced by SEDCNN-SVM, the important feature information is retained. When the feature information is input into SVM, the operation time and complexity are reduced, so the classification accuracy of sEMG is improved.

Fig. 10 shows the confusion matrix of different gestures recognized by the SEDCNN-SVM classification method. Labels 0-9 represent the ten different gestures shown in Figure 3. From the confusion matrix, the recognition accuracy of each gesture is different. The accuracy magnitude is given diagonally in the matrix. Most of the gestures have relatively high recognition accuracy. The misclassifications were concentrated in the "up" gesture, possibly by the signal generated in the sensors when the movement is not finished.

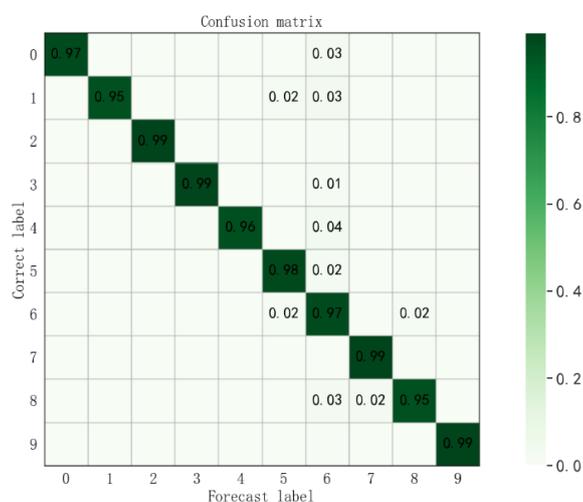

**Fig. 10.** Confusion matrix of ten gestures.

To prove the broad applicability of the algorithm, five healthy subjects are randomly selected for experimental evaluation. Table 5 shows the information about subjects, including three male subjects and two female subjects. SEMG data of five subjects are input into the classification model separately. The results of four measures used in Formulas (5) – (8) are obtained as shown in Figure. 11.

**Table 5.** Relevant information of different subjects.

| Subject | Gender | Age | Height[cm] | Weight[kg] |
|---|---|---|---|---|
| Subject1 | Male | 24 | 174 | 118 |
| Subject2 | Male | 25 | 176 | 138 |
| Subject3 | Male | 25 | 183 | 130 |
| Subject4 | Female | 26 | 168 | 110 |
| Subject5 | Female | 25 | 165 | 106 |

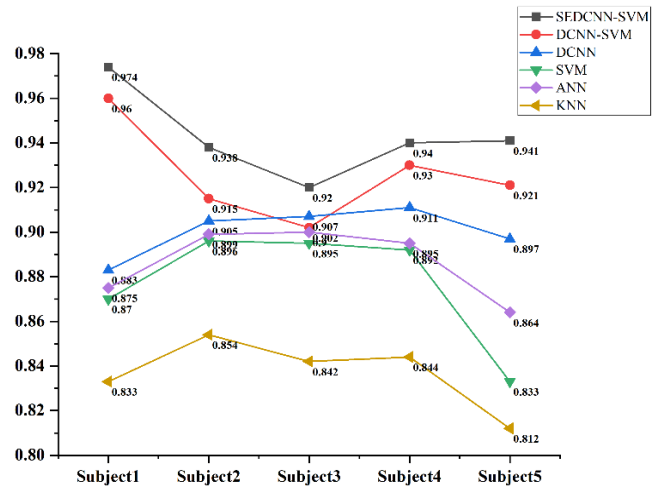

(a)

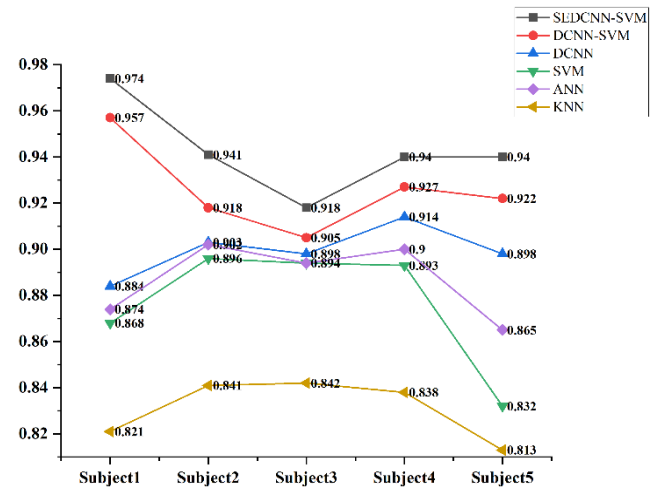

(b)

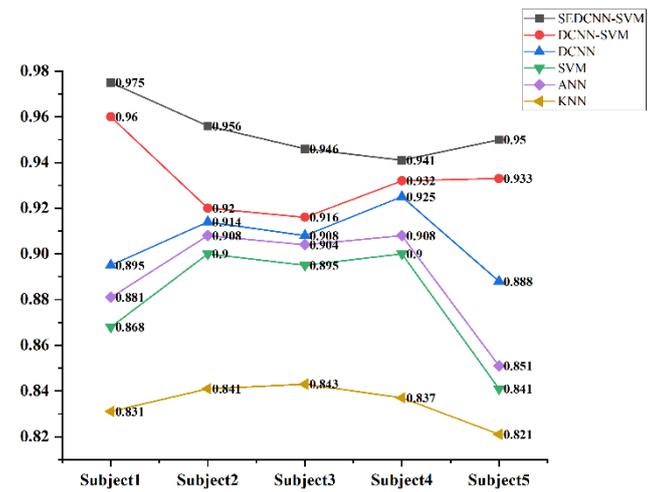

(c)

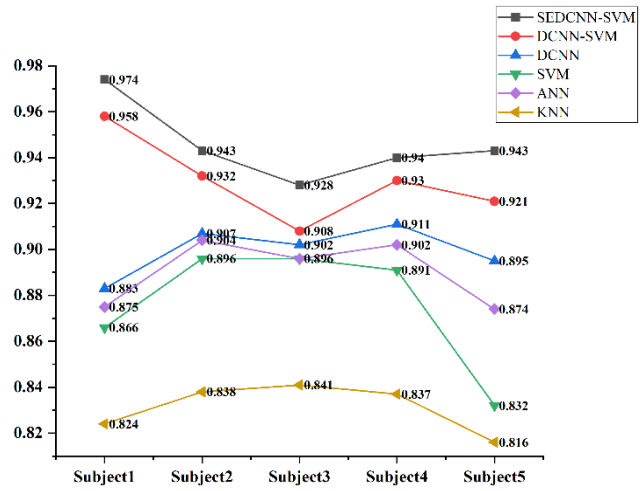

**(d)**

**Fig. 11.** Measurement indexes of different subjects under different classification models:(a) Accuracy (b) Recall (c) Precision (d) F1-score.

From the experimental results shown in the Figure 11, classification recognition methods have different accuracy in sEMG data sets of different subjects. This may be the result of different physiological conditions of the subjects themselves or different test environments. The recognition accuracy of the method proposed in this paper can reach more than 90% in each subject's data set.

For online prediction, the SEDCNN-SVM model is employed for gesture recognition tasks. One subject is randomly selected, the subject donned a Myo armband and executed six types of gestures on a manipulator grasping mode control platform. The results of online real-time gesture recognition are shown in Figure 12.

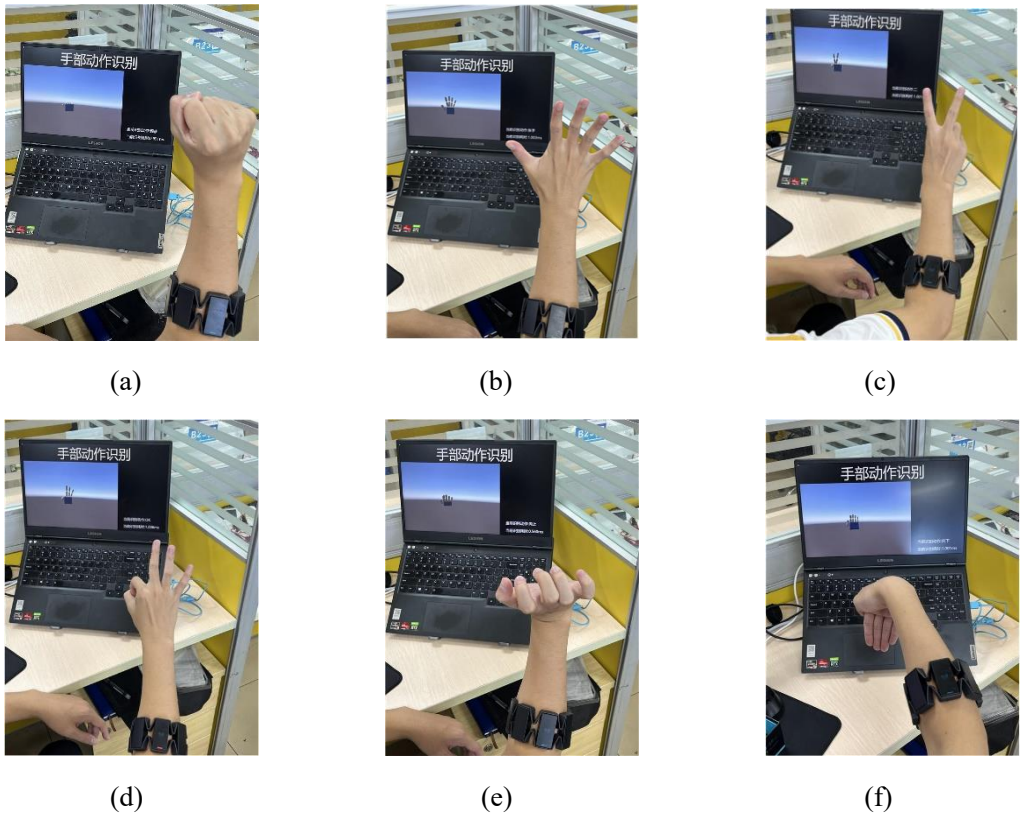

(a)                  (b)                  (c)

(d)                  (e)                  (f)

**Fig. 12.** Online gesture recognition: (a) fist (b) open hand (c) two (d) OK (e) up (f) down.

In the experiment of real-time recognition, 20 online recognitions were performed for each gesture respectively, a total of 80 experiments were performed. The experimental results are shown in Fig. 13. From the results in the figure, it can be concluded that the SEDCNN-SVM model can be well applied to the real-time recognition of hand gestures for each person, the number of correct classifications reaches 76 times in 80 experiments, so that the accuracy of correct recognition reaches 95%, with a high accuracy of real-time recognition.

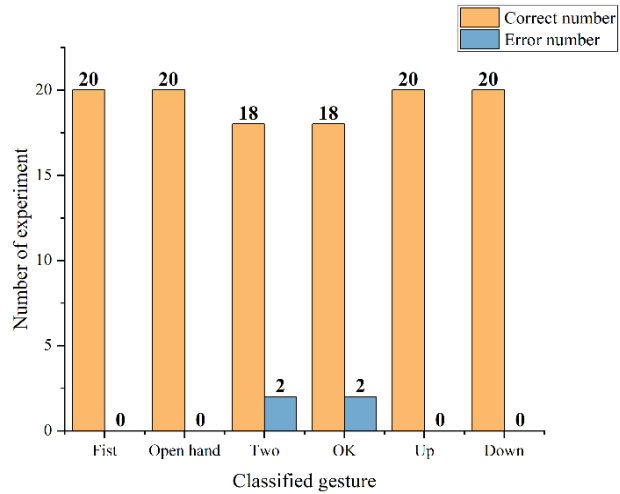

**Fig. 13.** The result of gesture recognition and classification in real time.

## 5. Conclusions

In this paper, an SEDCNN-SVM is proposed to recognize different gestures. To prevent the deterioration of classification performance caused by the deepening of convolutional neural network, SE-Net and residual modules are added to DCNN network. The feature vector of sEMG extracted from DCNN network was sent to SVM classifier for gesture recognition and classification. This method emphasizes the effectiveness of combining deep learning with machine learning and makes full use of their respective advantages and classification stability.

Experimental results show that the recognition accuracy of the SEDCNN-SVM classification algorithm is higher than that of the DCNN-SVM classification algorithm, and the recognition accuracy can be increased from 0.91 to 0.95, which fully demonstrates the effectiveness of SE-Net and residual module for feature extraction of DCNN network. By comparing the classification results of the last classification methods, it is found that the classification method of DCNN-SVM is superior to other classification methods. In the experiment of gesture real-time classification, SEDCNN-SVM model is proved to be suitable for different gesture real-time classification recognition.

For future research, different attention mechanisms can be added to deep convolutional neural network for comparison. The classification method can be applied to the classification of dynamic gestures. It provides a promising method for real-time control of rehabilitation mechanical arm band and other hardware equipment.

**Author Contributions**

Conceptualization, M.Z. and J.W.; methodology, M.Z.; software, M.Z.; validation, M.Z.; formal analysis, M.Z.; investigation, M.Z.; resources, M.Z. and J.W.; data curation, M.Z. and J.W.; writing—original draft preparation, Z.M. and J.W.; writing—review and editing, M.Z., J.W. and Q.W.; visualization, M.Z.; supervision, M.Z. and J.W.; project administration, M.Z., J.W., and Q.W.; funding acquisition, J.W. and Q.W. All authors have read and agreed to the published version of the manuscript

**Data Available**

The programs data used to support the findings of this study are available from the corresponding author upon request.

**Conflicts of Interest**

The author(s) declare(s) that there is no conflict of interest regarding the publication of this paper. We declare that we have no financial and personal relationships with other people or organizations that can inappropriately influence our work, there is no professional or other personal interest of any nature or kind in any product, service and/or company that could be construed as influencing the position presented in, or the review of, the manuscript entitled 'Estimation of finger joint angles with different gestures based on sEMG'.

**Funding Statement**

This work is supported by The National Natural Science Foundation of China (No.62072335)

**Ethical Compliance**

Research experiments conducted in this article with humans were approved by the Ethical Committee and responsible authorities of our research organization following all guidelines, regulations, legal, and ethical standards as required for humans.